\documentclass[journal]{IEEEtran}
\usepackage{amsfonts}
\usepackage{cite,graphicx,amsmath,amsthm}
\usepackage{subfigure}
\usepackage{fancyhdr}
\usepackage{dsfont}
\usepackage{array,color}
\usepackage{bm}
\usepackage{float}
\usepackage{algorithm}
\usepackage{algpseudocode}
\usepackage{multirow}
\usepackage{bbm}
\usepackage{graphicx}
\usepackage{float}
\usepackage{subfigure}
\usepackage{ragged2e}
\usepackage{booktabs}
\usepackage{bbding}
\usepackage{multirow}
\usepackage{makecell}
\usepackage{amsmath,amssymb,amsfonts}
\usepackage[]{caption2}
\usepackage[table,xcdraw]{xcolor}
\usepackage{cite}
\usepackage{hyperref}

\begin{document}

\title{\huge \textbf{Federated Deep Learning Meets\\Autonomous Vehicle Perception: Design and Verification}}

\author{Shuai Wang$^{*}$, Chengyang Li$^{*}$, Derrick Wing Kwan Ng,~\IEEEmembership{Fellow,~IEEE}, Yonina C. Eldar,~\IEEEmembership{Fellow,~IEEE},\\H. Vincent Poor,~\IEEEmembership{Life~Fellow,~IEEE}, Qi Hao, and Chengzhong Xu,~\IEEEmembership{Fellow,~IEEE}
\thanks{
\scriptsize
Shuai Wang is with Shenzhen Institute of Advanced Technology, Chinese Academy of Sciences.

Chengyang Li is with Hong Kong University of Science and Technology (Guangzhou).

Derrick~Wing~Kwan~Ng is with University of New South Wales.

Yonina C. Eldar is with Weizmann Institute of Science.

H.~Vincent~Poor is with Princeton University.

Qi Hao is with Southern University of Science and Technology and Shenzhen Research Institute of Trustworthy Autonomous Systems.

Chengzhong Xu is with IOTSC, University of Macau.

$*$: Equal contribution.
Corresponding author: Qi Hao and Chengzhong Xu.
}
}
\maketitle

\begin{abstract}
Realizing human-like perception is a challenge in open driving scenarios due to corner cases and visual occlusions.
To gather knowledge of rare and occluded instances, federated learning assisted connected autonomous vehicle (FLCAV) has been proposed, which leverages vehicular networks to establish federated deep neural networks (DNNs) from distributed data captured by vehicles and road sensors.
Without the need of data aggregation, FLCAV preserves privacy while reducing communication costs compared with conventional centralized learning.
However, it is challenging to determine the network resources and road sensor placements for multi-stage training with multi-modal datasets in multi-variant scenarios.
This article presents networking and training frameworks for FLCAV perception.
Multi-layer graph resource allocation and vehicle-road contrastive sensor placement are proposed to address the network management and sensor deployment problems, respectively.
We also develop CarlaFLCAV, a software platform that implements the above system and methods.
Experimental results confirm the superiority of the proposed techniques compared with various benchmarks.
\end{abstract}

\IEEEpeerreviewmaketitle

\section{Introduction}

\begin{figure*}[!t]
\centering
\subfigure[]{\includegraphics[width=1\textwidth]{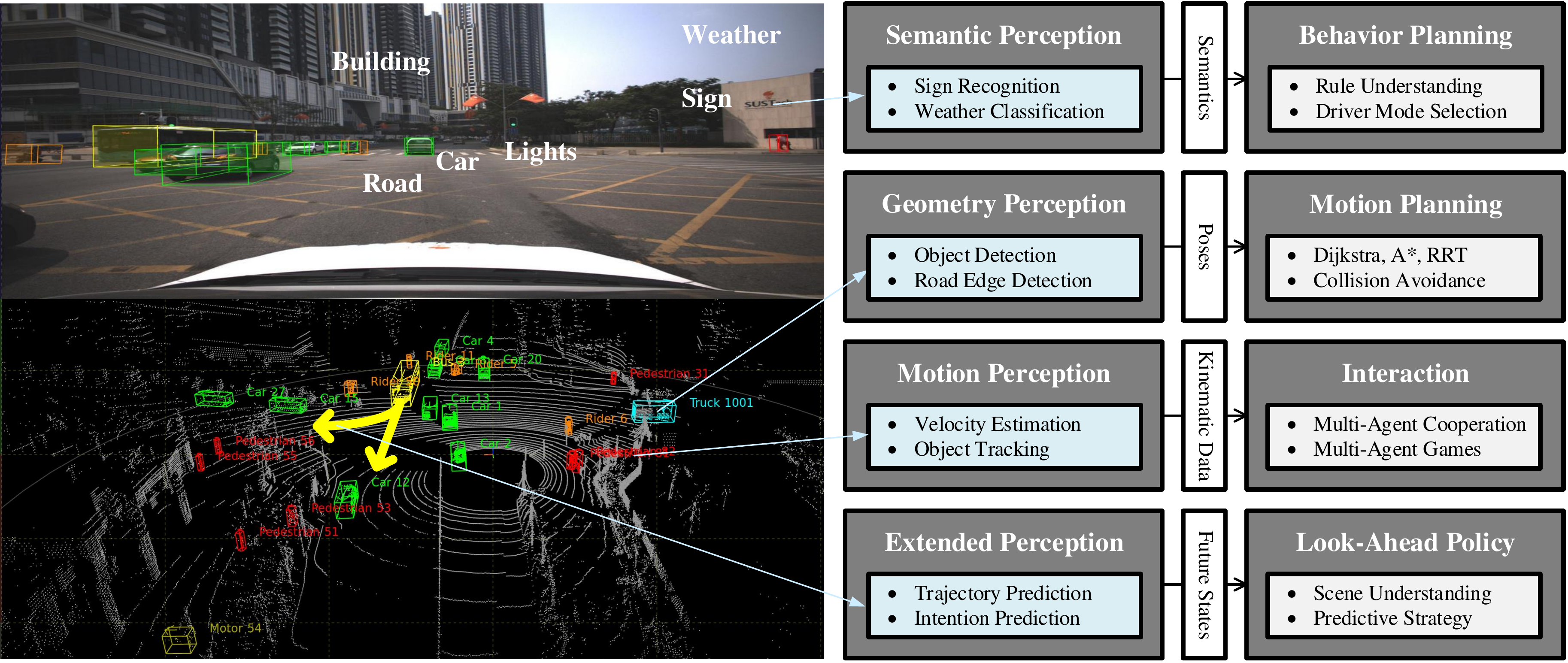}}
\subfigure[]{\includegraphics[width=1\textwidth]{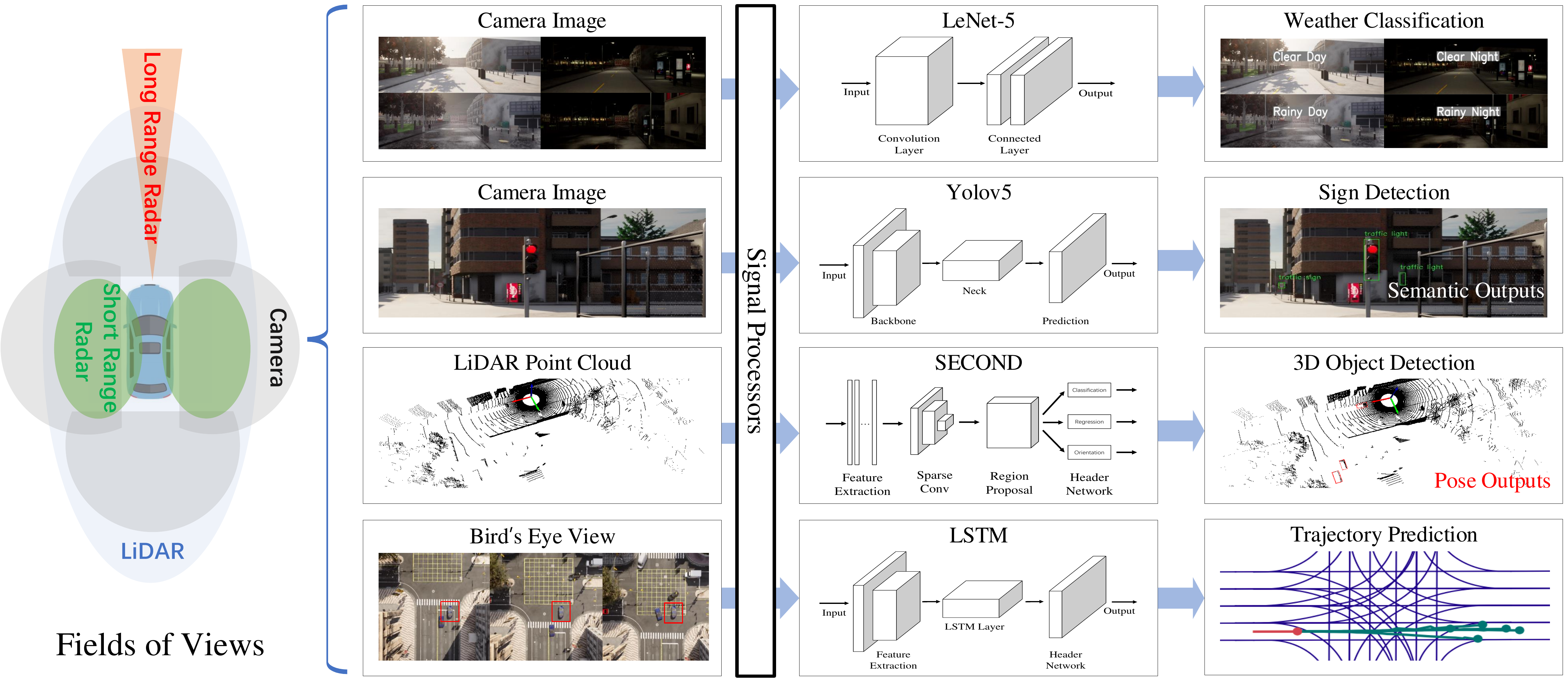}}
\caption{a) Categories of CAV perception.
b) Multi-task multi-modal deep learning for CAV perception.
From top to bottom: 1) weather classification (task 1); 2) sign recognition (task 2); 3) object detection (task 3); 4) trajectory prediction (task 4).
From left to right: 1) sensors map the environment into digits (raw data) using active or passive electromagnetic waves; 2) signal processors clean, rotate, crop, or complete the raw data; 3) DNNs extract features and generate semantic, geometry, motion, and prediction outputs. All datasets are generated by the CarlaFLCAV platform.}
\end{figure*}

Perception determines the way a connected autonomous vehicle (CAV) understands the world by transforming environments into digits via sensors, signal processors, and deep neural networks (DNNs) \cite{perception}.
Conventionally, DNN training is based on centralized learning, which collects the datasets from CAVs, trains the DNNs at the cloud, and deploys trained DNNs on CAVs for inference.
This paradigm is effective in closed driving areas where the owner of data-collection vehicles is also the DNN provider.
However, future CAV systems must cope with multi-variate open scenarios, which involves corner cases due to infinite scenario space and visual occlusions due to high scenario complexity \cite{perception}.
CAV companies, e.g., Tesla, Waymo, Baidu, suggested that these challenges be tackled via lifelong multi-stage training that updates the DNN parameters whenever rare or occluded objects are detected (e.g., Tesla Autopilot and over-the-air updates solution \href{https://karpathy.ai/}{https://karpathy.ai/}).
In this case, sensor data is inherently distributed at customers' vehicles and contains high-resolution human-related private information, leading to the potential of sensitive information leakage \cite{flcav_app1}.

Federated (deep) learning assisted CAV (FLCAV) is an emerging paradigm to overcome the privacy issue by training DNNs via parameter and output aggregation instead of dataset aggregation \cite{flcav_app1,flcav_app2,flcav_netw1,flcav_netw2,flcav_road1,flcav_road2}: parameter aggregation leverages vehicular networks to migrate knowledge among different vehicles \cite{flcav_netw1,flcav_netw2}; output aggregation leverages road sensors' perception results to annotate occluded objects for local parameter updates \cite{flcav_road1,flcav_road2}.
The performance of FLCAV highly depends on the associated network resources and road sensor placements.
{\color{black}Existing works on FLCAV design these factors from either a driving (e.g., \cite{flcav_app1,flcav_app2}) or networking (e.g., \cite{flcav_netw1,flcav_netw2}) perspective, which ignores the interdependency between driving tasks (i.e., data consumer) and vehicular networks (i.e., data provider).
This research gap has been identified as the core issue in CAV systems \cite{cav_netw}.
Yet, how to close this gap for FLCAV is still an open issue.

This article integrates driving and networking features for \emph{system-level} FLCAV perception under practical network resource constraints.}
Frameworks of vehicle-edge-cloud networking, multi-stage DNN training, and multi-task generation, are presented.
On top of these frameworks, multi-layer graph resource allocation (MLGRA) and vehicle-road contrastive sensor placement (VRCSP) are proposed. The MLGRA jointly allocates the limited network resources across different stages, tasks, and modalities to minimize perception errors. The VRCSP automatically reduces (increases) the number of sensors in low (high) complexity scenarios. To verify the above methods, it is necessary to implement a simulator with high-fidelity rendering qualities, driving behaviors, and software interfaces \cite{aads}, as FLCAV needs long-term training that cannot be quickly tested in reality.
However, currently there is no such close-to-reality FLCAV simulator.
We thus develop CarlaFLCAV (\href{https://github.com/SIAT-INVS/CarlaFLCAV}{https://github.com/SIAT-INVS/CarlaFLCAV}), an open-source software platform for design and verification of FLCAV systems. The platform contains dataset generation, perception tasks, FL frameworks, and optimization modules, and aims to provide a concrete step towards FLCAV in the real world.

\section{FL Meets CAV}

\subsection{CAV Perception: Concept, Challenges, and Trends}

\subsubsection{CAV Perception}

As shown in Fig.~1a, a CAV perception system is an ensemble of the following functionalities \cite{perception}:
1) semantic perception, e.g., recognition of lanes, road arrows, and traffic signs, which outputs semantics for rule understanding;
2) geometry perception, e.g., road edge detection and object detection, which outputs poses (including positions, sizes, and orientations) for collision-avoidance motion planning;
3) motion perception, e.g., object tracking, which outputs kinematic data (velocities) for multi-agent interaction.
4) extended perception, e.g., trajectory prediction, which outputs future states of agents for look-ahead decision making.
As shown in Fig.~1b, each functionality is decomposed into a set of heterogenous tasks, calling for multi-modal sensors such as RGB/infrared camera, LiDAR, radar, GPS and IMU to exploit their complementary features, e.g., field of views (FoVs). Each task is accomplished by sensors, processors, and DNNs. For instance, in trajectory prediction, images generated by multi-view cameras are fed to vision transformers, generating bird's-eye-view data sequences that are fed to the long short-term memory (LSTM) module to forecast what the observed traffic agents shall do.
By observing the past coordinates of a vehicle (marked in red box), this pipeline successfully predicts future coordinates of a vehicle moving through the cross-road.

\subsubsection{Perception Challenges}
Major challenges of CAV perception in multi-variate open scenarios are summarized below.
\begin{itemize}
\item
\textbf{Corner Case}. Scenario space in open areas grows exponentially, making it impossible to enumerate all the possible cases during the training stage \cite{perception}.
In other words, there always exists new data outside the distribution of training datasets, which are corner case instances.

\item
\textbf{Visual Occlusion}.
In complex urban scenarios (e.g., cross-road, T-junction, roundabout), an object can be occluded by another in the FoV, which leads to incomplete data and perception errors \cite{flcav_road1}.

\item
\textbf{Verification Cost}.
Releasing new DNNs requires rigorous verification.
The cost of real-word testing is not acceptable due to the long-term training required by FLCAV and difficulties of testing in dangerous scenarios (e.g., crashes, overtaking, bad weather) \cite{carla}.
\end{itemize}

\subsubsection{Research Trends}

Possible solutions to address the above challenges are as follows.
\begin{itemize}
\item
\textbf{Lifelong multi-stage training}, which updates the parameters whenever a corner case is detected, is a promising solution to address the first challenge \cite{flcav_app2}.
The update is executed on DNN copies, making sure that the inference DNNs are fixed during the training process.
Releasing new-version DNNs requires the acknowledgement of DNN providers and customers.
\item
As for the occlusion issue, the solution is \textbf{cooperative perception with road sensors}.
Practical road sensors are authoritative, having absolute positions, broader FoVs, and highly-optimized hardware units \cite{flcav_road1}.
Furthermore, the probability of a target object being occluded in FoVs of all road sensors is significantly smaller than that in the FoV of a CAV.
As such, the perception error due to occlusions is significantly mitigated.
\item
To reduce the high verification cost in reality, \textbf{CAV simulation} becomes a necessity \cite{aads}.
Various CAV simulators have been released, e.g., Intel Car-Learning-to-Act (CARLA), Nvidia DRIVE, Microsoft Aerial-Informatics-and-Robotics-Simulation (AirSim), LG Silicon-Valley-Lab (LGSVL), Baidu Augmented-Autonomous-Driving-Simulation (AADS), UCLA OpenCDA, SJTU V2X-Sim \cite{flcav_road2,aads,carla}.
\end{itemize}

\subsection{Paradigm Shift: From Centralized to Federated Learning}

\subsubsection{Federated Learning}

The need for multi-stage training results in the shift of learning paradigms.
In one-stage training, the owner of data-collection vehicles is also the DNN provider and the aggregated datasets can be directly fed to DNN training pipelines.
However, for multi-stage training, the abnormal data is distributed at customers' vehicles and contains human-related private information.
As a consequence, conventional centralized learning is no longer applicable.
To this end, FLCAV emerges, which trains the DNNs from distributed datasets via parameter aggregation, thus conveying the knowledge of corner cases to other vehicles and remote servers while preserving data privacy \cite{flcav_app1}.
In addition, communication costs are reduced, since the size of a DNN is significantly smaller than that of a data sequence, e.g., the size of a 1-minute point-cloud sequence is $1000\,$MB while that of a standard object detection DNN is $60\,$MB \cite{perception} (FL takes multiple rounds of communication; but with proper pre-training, only a few rounds are needed).

\subsubsection{Federated Distillation}

FL can be integrated with cooperative perception, giving rise to the federated distillation (FD) technique \cite{flcav_road2}.
In the FDCAV framework, all road sensors and CAVs upload their bounding boxes and perception uncertainties to a road server.
The server computes the weighted average of these outputs and this road-average output is downloaded to surrounding vehicles.
Each vehicle updates its local parameters by minimizing the contrastive loss between its bounding boxes and the road-average boxes \cite{flcav_road2}.
Note that there exist non-negligible computation and communication delays during the vehicle-to-infrastructure output fusion progress.
However, FDCAV is a long-term DNN-update framework rather than a real-time detection method for collision avoidance.
Therefore, it is possible to associate data from the roadside infrastructure with that from CAVs by fusing a sequence of data frames via pose graph optimization, which removes the effects of random errors by extracting multi-frame spatial-temporal patterns.
FDCAV can also adopt spatial contexts of road maps for generating ground-truth semantics and training associated DNNs.
For instance, the Road Experience Management (REM) of Mobileye (\href{https://www.mobileye.com/our-technology/rem/}{https://www.mobileye.com/our-technology/rem/}) adopts road sensors and maps to realize automated road semantic identification and annotation.

\begin{figure*}[!t]
\centering
\subfigure[]{\includegraphics[width=1\textwidth]{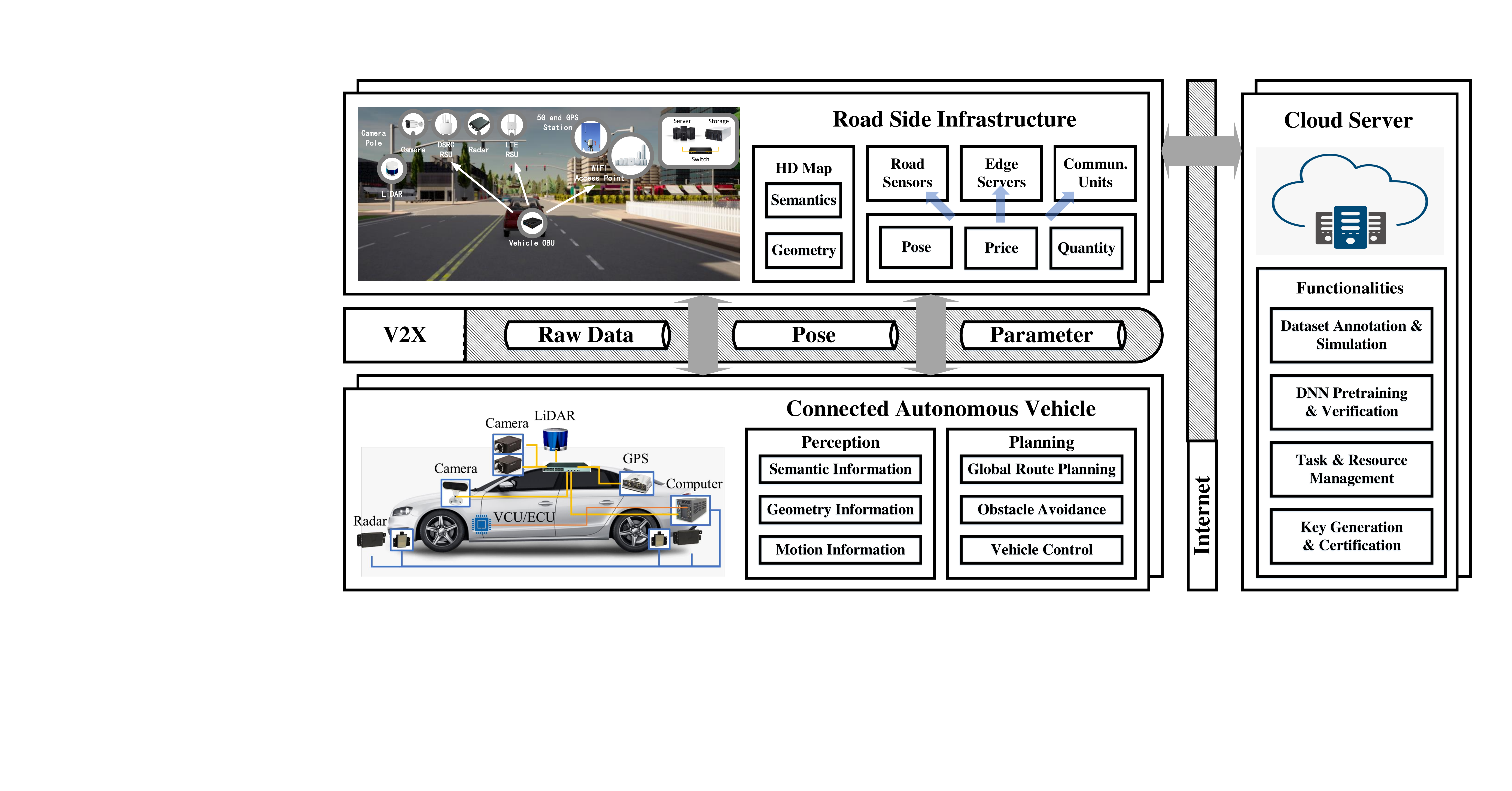}}
\subfigure[]{\includegraphics[width=1\textwidth]{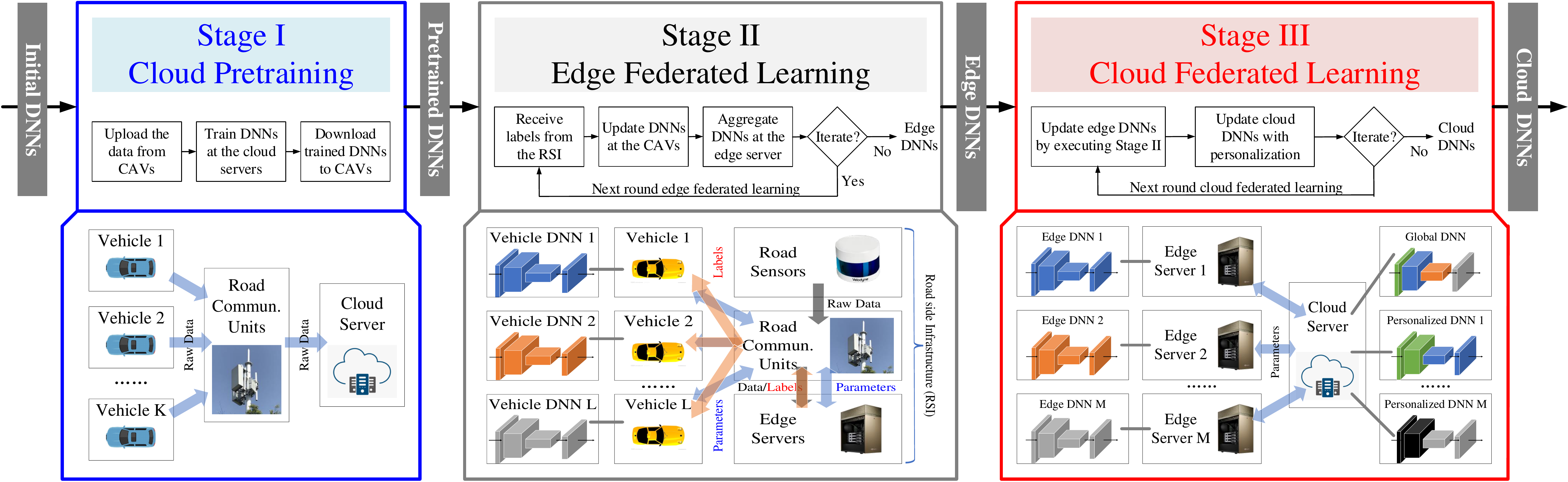}}
\caption{{\color{black}a) Network architecture of FLCAV with vehicle, road, and cloud components and their associated V2X and the Internet links; b) Multi-stage training pipeline of FLCAV with intra-stage and inter-stage network flows.}
}
\end{figure*}

\subsection{Related Work on FLCAV}

\subsubsection{Limitations of Existing Work}
Current literature on FLCAV can be categorized into two types:
1) network-layer designs (e.g., FL resource allocation), and 2) application-layer designs (e.g., FL perception and planning).
Specifically, network-layer designs, e.g., \cite{flcav_netw1,flcav_netw2,flres1}, aim to improve FL performance by controlling the communication-related variables such as topology, throughput, latency, and device scheduling. These works address the challenges of FL from the network perspective while taking the high mobility of vehicles into account.
However, they fail in matching FL to domain-specific CAV scenarios, tasks, datasets, and DNNs.
On the other hand, application-layer designs, e.g., \cite{flcav_app1,flcav_app2,flcav_road1,flcav_road2}, improve the safety and efficacy of CAV systems by designing DNN structures and associated information fusion methods.
These works provide solutions to CAV issues such as training, inference, synchronization, calibration, and simulation.
Nonetheless, communications among CAVs, edge servers, and cloud servers therein are assumed to be perfect, which does not hold for practical CAV systems with limited resources.

\subsubsection{Research Opportunities}
Since both types of designs have different pros and cons, it is necessary to integrate them to achieve lower detection/classification/tracking errors under network resource and sensor implementation constraints.
This leads to new technical problems, which cannot be tackled by conventional methods.
\begin{itemize}
\item
\textbf{Opportunity 1 (Network Resource Allocation)}: How can we effectively distribute the network resources across different stages, tasks, and modalities to minimize the perception errors of final-stage DNNs while satisfying the stringent wireless and wireline communication constraints? Existing literature does not analyze CAV-domain-specific datasets and ignores the interdependency across different stages.

\item
\textbf{Opportunity 2 (Sensor Placement)}: How can we efficiently place the road sensors to detect and annotate more occluded objects under the implementation cost constraint?
Conventional approaches adopt integer programming (IP) solvers or heuristic methods to maximize the coverage with a fixed number of sensors, which ignores the learning requirements for FLCAV.

\item
\textbf{Opportunity 3 (Software Engineering)}: How can we implement a high-fidelity FLCAV simulator that is close to reality?
Existing methods of network-layer designs are tested in simple classification tasks (e.g., recognition of handwritten digits).
Emerging autonomous driving simulators do not support FL and associated optimization functionalities.
\end{itemize}

\section{System-Level Design for FLCAV Perception}

\subsection{Network Architecture}

As shown in Fig.~2a, FLCAV consists of CAV, road, and cloud components, forming a wide-range cyber-physical system.
CAVs are integrated systems equipped with multi-modal sensors, mobile computing platforms, and advanced communication units that 1) connect the vehicle sensors, computers, and bases via controller area network (CAN), and 2) connect the vehicles with the cloud and road via vehicle-to-everything (V2X) technology \cite{cav}.
CAV is a server within its CAN and a client within its V2X network.

{\color{black}
Roadside infrastructure can be categorized into road sensors, road communication units, and road computing servers \cite{cav_netw}.
First, the pose of a road sensor includes position and yaw/roll/pitch angle, which directly determines the coverage of the environment and objects therein.
Second, road communication units adopt the V2X to link surrounding vehicles, forming a vehicular edge network.
Finally, road servers are parameter servers for edge FL and process real-time tasks, e.g., feeding the data of road sensors into road DNNs, merging the multi-sensor data using fusion techniques (e.g., iterative closest point (ICP)).

A cloud server differs from a road (edge) server since the Internet is a part of the end-to-end communication.
Therefore, cloud servers execute long-term tasks, e.g., dataset annotation, DNN training, simulation, and resource management.
Note that to prevent potential cyber attacks, any vehicle joining FLCAV should be registered with a unique identifier allocated by authorities.
The cloud server is responsible for maintaining the identifier and also executes security key generation and certification \cite{cav_netw}.

\subsection{Training Pipeline}

Training FLCAV perception systems in Fig.~2b consists of cloud pretraining, edge FL, and cloud FL stages.
Specifically, cloud pretraining adopts annotated datasets on the cloud to transform initial DNNs into pretrained DNNs that are released to all CAVs.
Then, an FL request is generated at the CAV upon a false detection event (e.g., due to occlusions or corner cases).
The FL request is sent to the edge parameter sever, who calls for roadside infrastructures and other vehicles to join the edge FL group via V2X.
This helps the CAV fix the bug residing in its local DNN, as the knowledge migrates from other agents to it and vice versa.
With a few rounds of output and parameter exchange, the edge parameter server can form edge DNNs that serve as good representations of the local region.
Finally, the FL group can be enlarged by cloud FL, which aggregates the edge DNNs from remote areas via the Internet.
To improve robustness, the cloud FL stage may adopt personalization such that each edge client trains its own regional DNNs while contributing to the global cloud DNNs \cite{personalization}.

From the application perspective, FLCAV needs to train a set (let $N$ denote its cardinality) of DNNs for associated perception tasks in 3 (or more) consecutive stages.
From the network perspective, FLCAV involves two types of communications, i.e., wireless and wireline communication, and their transmission capacities are finite.
Combining both, the summation of wireless/wireline throughput (in MBytes) over $N$ tasks and 3 stages should be smaller than the network throughput budget, and there exists a tradeoff among different tasks and stages.
Here we consider the uplink transmission in Fig.~2b, as the downlink counterpart is usually not the bottleneck in practice.
\begin{itemize}
\item[1)] For Stage I, data samples should be uploaded from vehicles to the edge and then to the cloud.
The product of the number of samples and the data size of each sample should not exceed the minimum throughput of wireless and wireline communication allocated to Stage I.

\item[2)] For Stage II, parameters are exchanged between the edge server and vehicles through wireless communication.
The product of the number of edge FL rounds, the number of vehicles in each FL group, and the data size of DNNs should not exceed the wireless throughput allocated to Stage II.

\item[3)] For Stage III, parameters are exchanged between the edge and cloud servers through wireline communication.
The product of the number of cloud FL rounds, the number of edge servers, and the data size of DNNs should not exceed the wireline throughput allocated to Stage III.
\end{itemize}
Note that perception outputs (e.g., bounding boxes) are also shared among nodes, but the associated communication overhead is negligible compared with those of samples and DNNs.

\subsection{Task Generation}

Tasks should match scenarios \cite{task} and their construction shown in Fig.~3a consists of the following 4 steps.
\begin{itemize}
\item[1)]
\textbf{Operational Design Domain (ODD) Specification}. Given the target ODD, e.g., urban (focus of this paper), rural, campus, mine, port, or parking-lot, we define the key parameters including traffic density, speed limits, rules, and FoV requirements \cite{scenarios}.

\item[2)]
\textbf{Scenario Sampling}.
Scenarios are sampled inside the ODD according to industrial standards such as ISO and SAE \cite{scenarios}.
Learning and optimization based methods can also be adopted for scenario space exploration.
CarlaFLCAV provides straight-road, cross-road, T-road, and roundabout scenarios.

\item[3)]
\textbf{Task Generation}. Scenario-specific tasks are generated, where each task is defined as a set of data, labels and DNNs \cite{task}.
We construct fewer tasks for low-complexity scenarios to save computational costs and redundant tasks for high-complexity scenarios to guarantee driving safety.

\item[4)]
\textbf{Task Evaluation}.
Perception tasks are categorized into different priorities via a task importance evaluator, which removes a task from the task list and observes the associated performance loss. A task is deemed important if the performance loss is significant \cite{task}.
\end{itemize}
}

\section{Resource Optimization for FLCAV Perception}

\subsection{Multi-Modal Resource Allocation}

\begin{figure*}[!t]
\centering
\subfigure[]{ \includegraphics[width=1\textwidth]{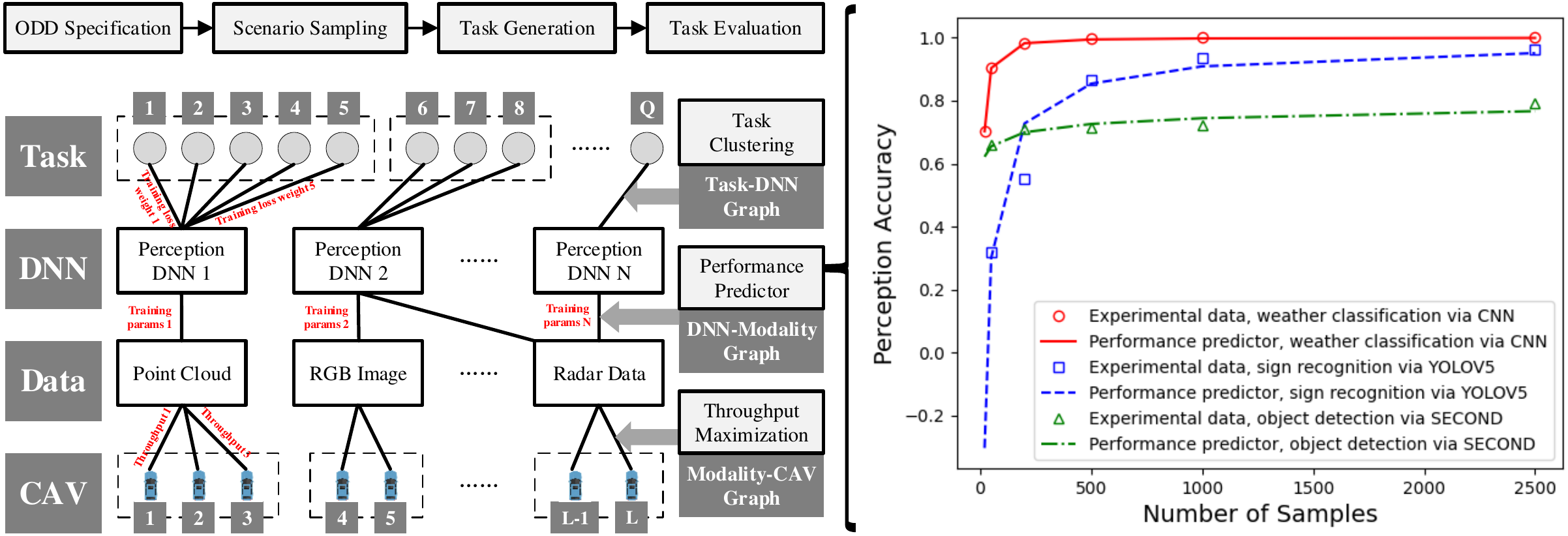}}
\subfigure[]{ \includegraphics[width=1\textwidth]{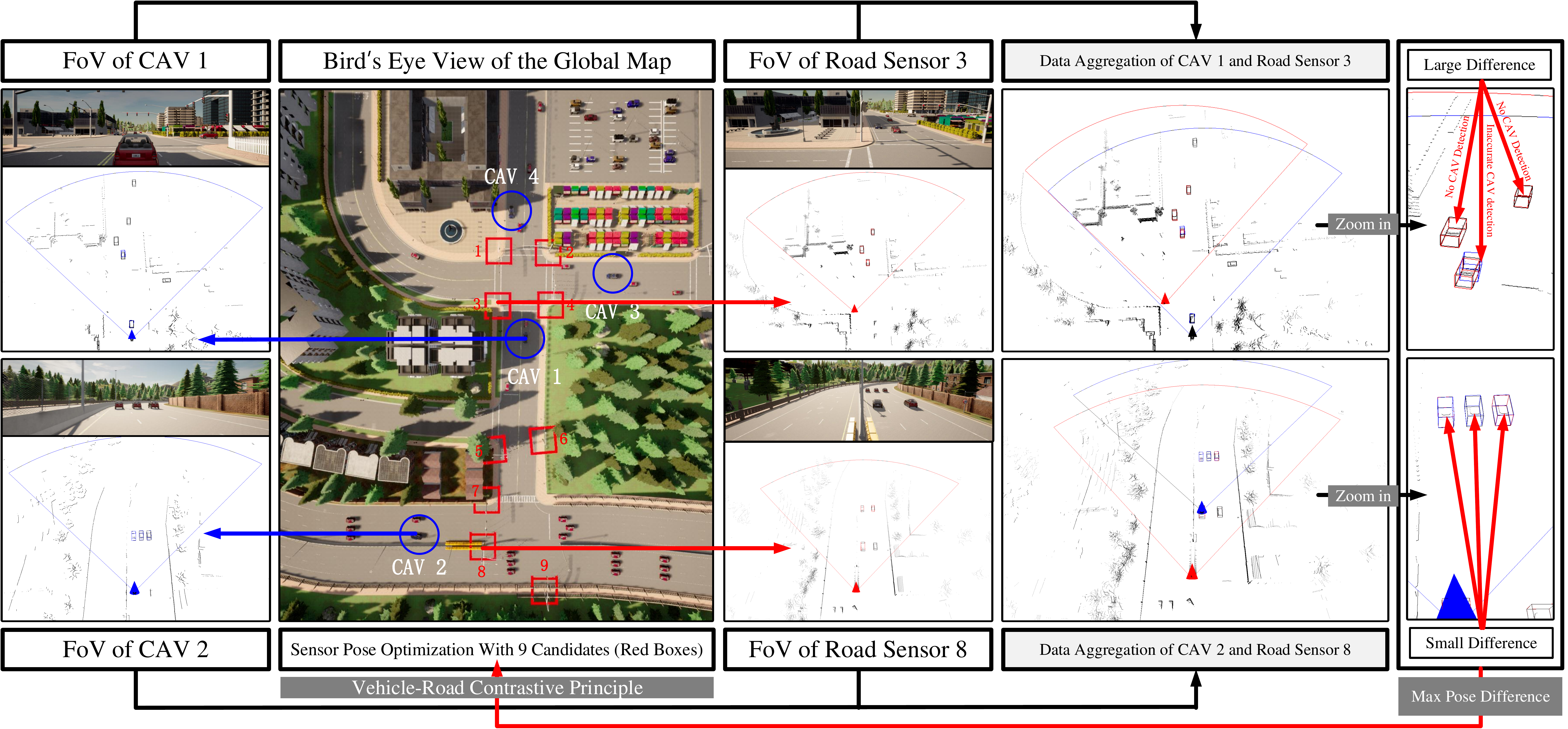}}
\caption{{\color{black}a) Illustration of the MLGRA method and the good fitness of the performance predictor to the experimental data generated by CarlaFLCAV; b) Illustration of the road-assisted FD and the VRCSP. The black box is the ground truth; the blue box is from the CAVs; the red box is from the road sensors.}
}
\end{figure*}

\subsubsection{Motivation}

Intuitively, more resources should be allocated to perception tasks accomplished by deeper neural networks.
Consider training a convolution neural network (CNN) for weather classification and a sparsely embedded convolutional detection (SECOND) network for object detection.
Due to larger number of parameters ($\sim$5 millions) in SECOND, the FLCAV network should allocate more communication resources to vehicles that upload point clouds for pretraining SECOND in Stage I, and call for more vehicles and FL rounds for sharing SECOND parameters in Stages II and III.
In current vehicular networks, the purpose of resource optimization is to improve key communication metrics, which treats data equally and violates the above intuition.

\subsubsection{Method}

Since communication (i.e., a data pipeline) becomes a sub-task in the FLCAV paradigm, we need to directly minimize the perception error instead of maximizing the communication throughput.
The problem becomes how to model the relation between perception errors and network flows.
Here we present a multi-layer graph resource allocation (MLGRA) approach shown in Fig.~3a.
For layer 1, vertex represents task or DNN, while edge represents training weight;
for layer 2, vertex represents DNN or data modality, while edge represents training parameters (e.g., number of training samples, number of FL rounds);
for layer 3, vertex represents data or CAV, while edge represents vehicle throughput.
\begin{itemize}
\item
\textbf{Task-DNN Graph}.
Tasks are clustered into groups such that similar tasks share the same DNN and annotated dataset.
For example, the tasks of box regression (i.e., determining the poses of objects), object classification (e.g., determining if the object is a car or a truck), and orientation classification (e.g., determining the front side) can be accomplished by one DNN with a common feature extractor and 3 separate headers.
The training loss function is the weighted sum of three metrics.

\item
\textbf{DNN-Modality Graph}.
Different DNNs may be trained with different data modalities, and their connections form a DNN-modality graph.
The required resource for a DNN-modality pair can be obtained by fitting performance predictors to historical datasets.
For instance, inverse-power models are adopted to predict perception accuracies of 3 DNN-modality pairs (i.e., CNN-Image, YOLOV5-Image, SECOND-PointCloud) under different communication resources in Fig.~3a.
The predictions match the experiment data very well for all DNNs.

\item
\textbf{Modality-Vehicle Graph}.
For each data modality (e.g., point cloud data), the associated data samples may come from multiple CAVs.
If the data is independent and identically distributed (IID) among different vehicles, we can maximize the total throughout via some optimization algorithm.
If the data is non-IID due to different FoVs, it is necessary to evaluate the quality of vehicles' data.
\end{itemize}

\subsection{Road Sensor Placement}

\begin{figure}[!t]
\centering
\includegraphics[width=0.5\textwidth]{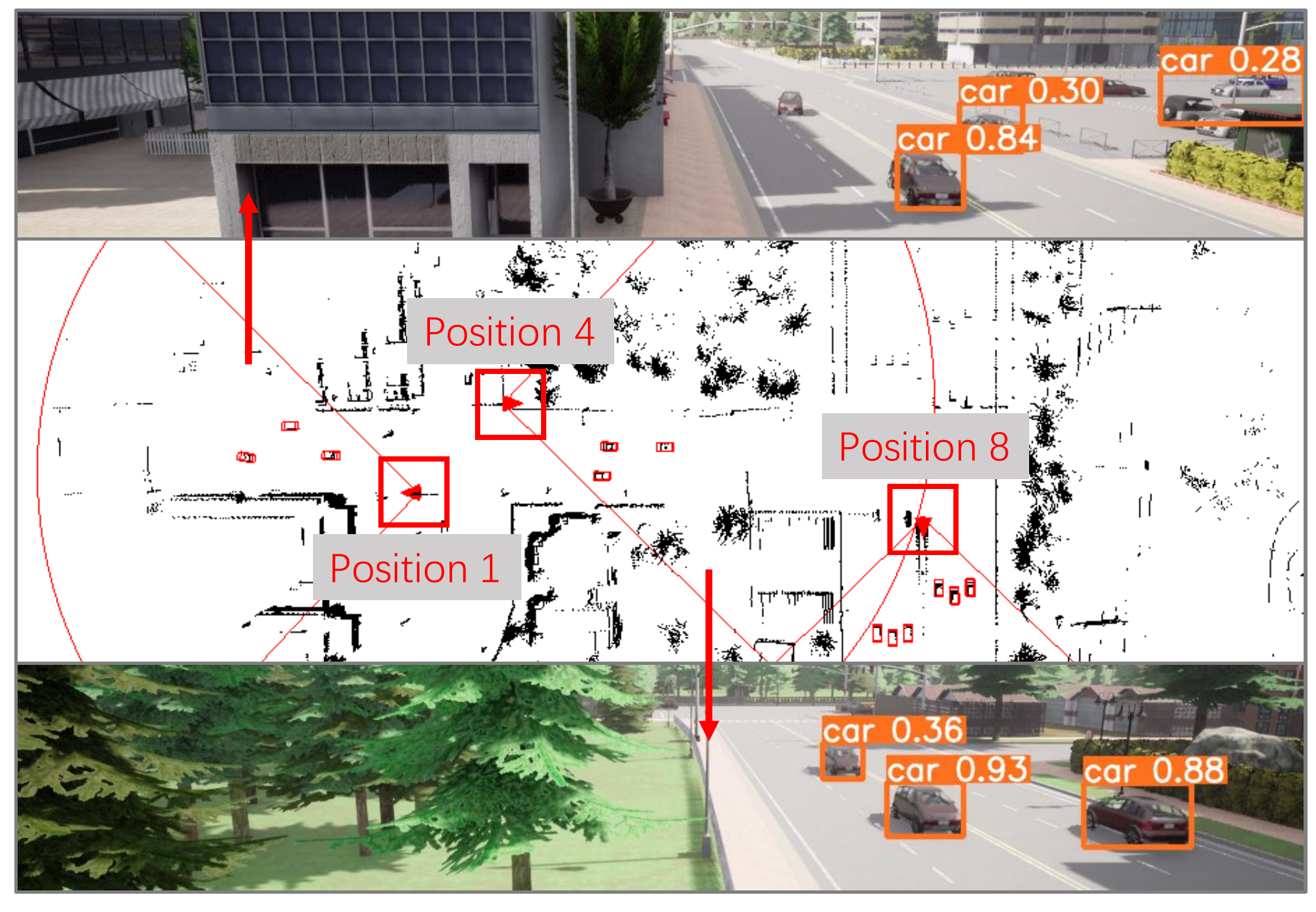}
\caption{
Illustration of multi-modal multi-view sensor fusion at the roadside infrastructure.}
\end{figure}

\begin{figure*}[!t]
\centering
\subfigure[]{\includegraphics[height=0.37\textwidth]{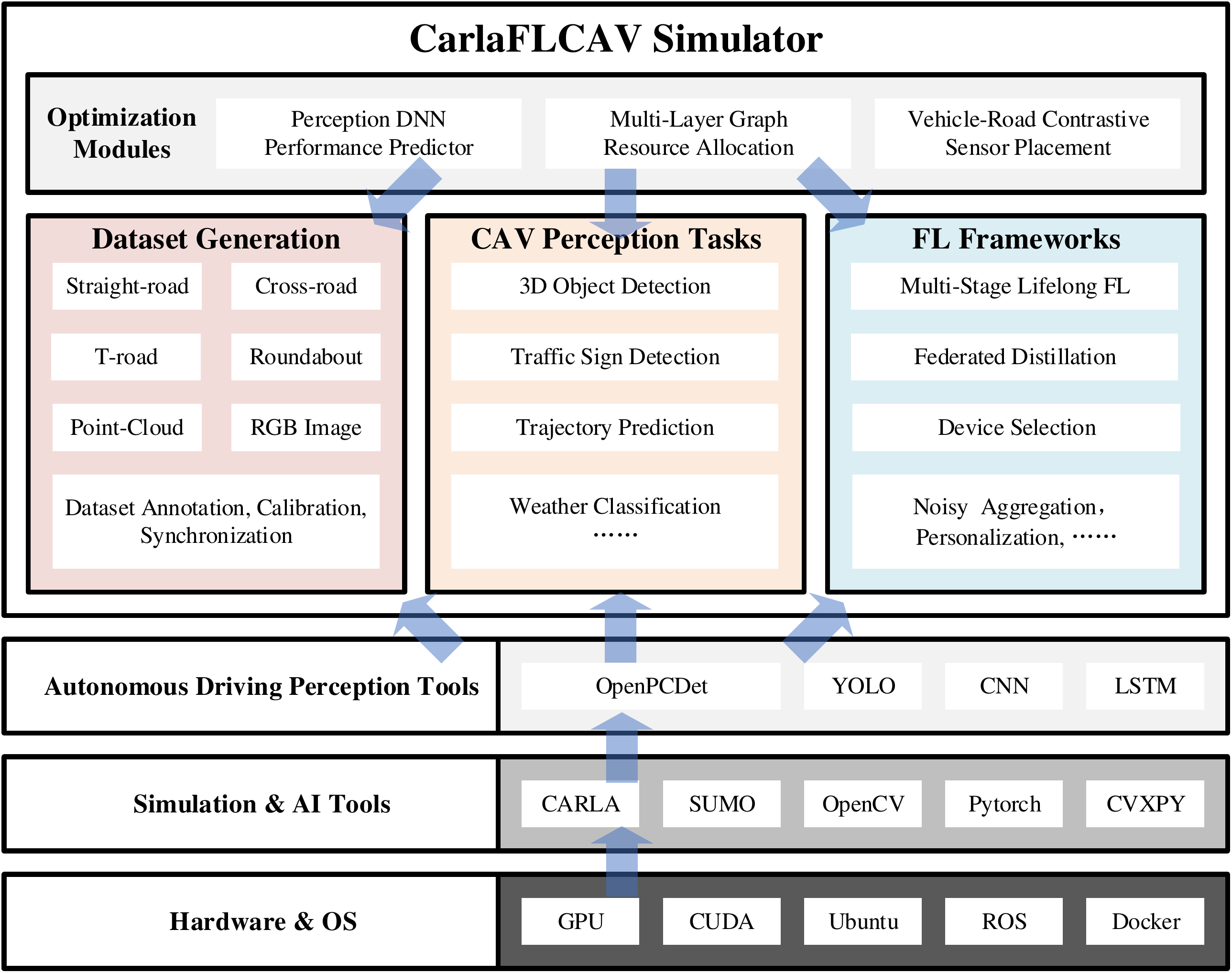}}
\subfigure[]{\includegraphics[height=0.37\textwidth]{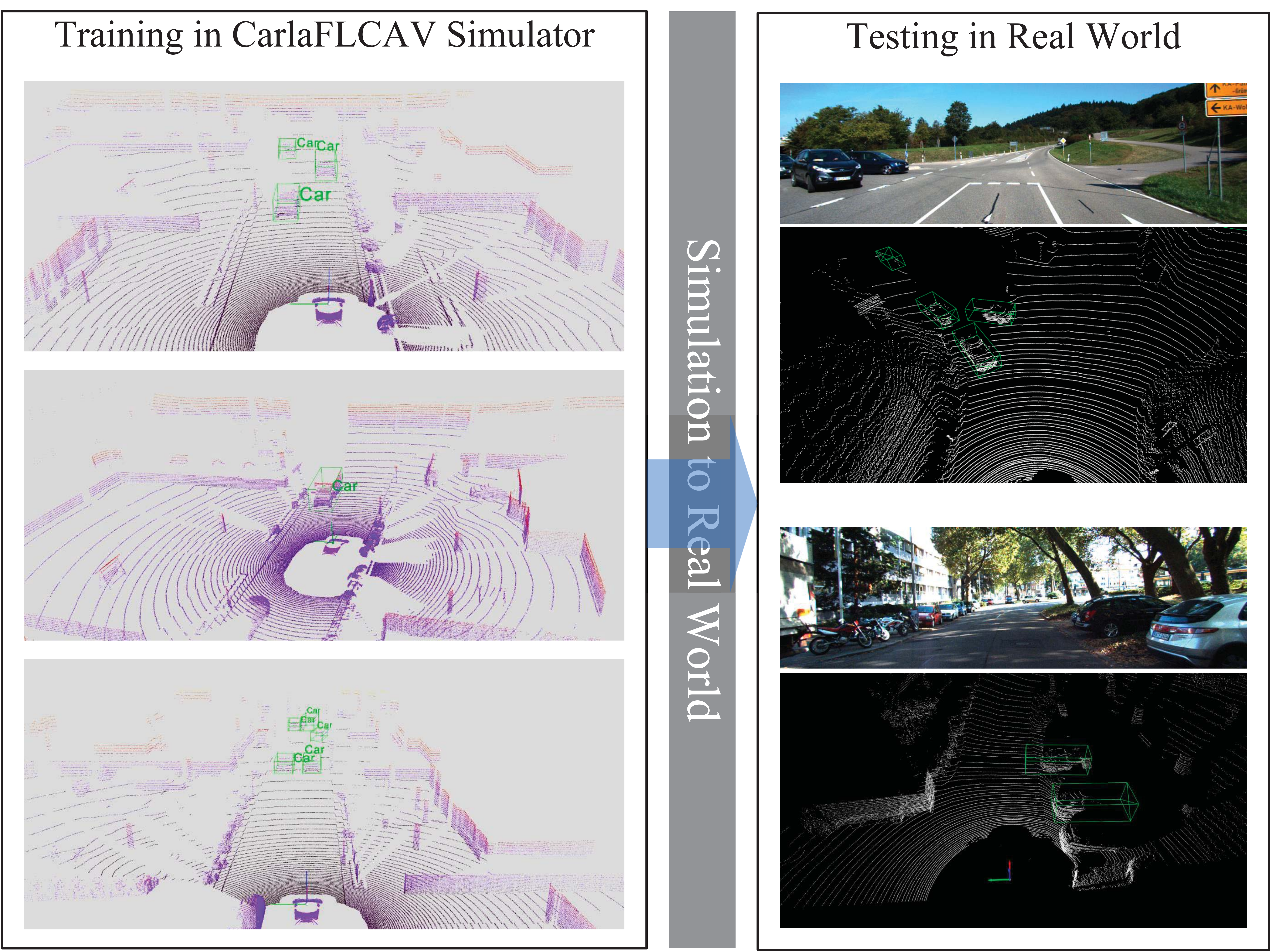}}
\caption{{\color{black}a) Software architecture and supported functionalities of CarlaFLCAV; b) Training DNNs in the CarlaFLCAV simulator and testing with real world datasets.
The DNN trained with CarlaFLCAV successfully detects all objects in the real world.}
}
\end{figure*}

\subsubsection{Motivation}

For FLCAV, the key is to upgrade the onboard DNNs instead of monitoring the environment.
Therefore, in contrast to conventional methods that maximize the number of visible objects \cite{flcav_road1}, we place sensors at critical scenarios that contain adversarial objects that can defeat the DNNs.
To illustrate their difference, we adopt CarlaFLCAV to generate an urban traffic map with straight-road and cross-road scenarios, where 9 possible sensor locations are marked as red boxes in the middle of Fig.~3b.
\begin{itemize}
\item Conventional methods \cite{flcav_road1} place the road sensor suite (i.e., consisting of a 360-degree LiDAR and an RGB-camera) at location 8, as we set its nearby traffic density to the highest value. However, the vehicle DNN would not learn any new knowledge, as the bounding boxes generated by the vehicle (e.g., CAV 2) and the road sensor 8 are similar.

\item Our method places the road sensor suite at location 3, despite the fact that its nearby traffic density is low. Here, the CAV could change its local parameters to the maximum extent with the road's outputs.
This is because the bounding boxes generated by the vehicle and road sensor 3 are very different.
For example, CAV 1 misses two objects since its FoV is occluded by its front car.
\end{itemize}

\subsubsection{Method}

To find the critical scenarios, our VRCSP minimizes the pose similarity between the detected objects at the CAV and those at the road sensor.
In particular, we deploy the pretrained DNN on a CAV and test the vehicle in a target map containing multiple scenarios, each specified with zone limits.
If the CAV fails to detect an object or generates a false positive in its FoV, the 3D position of this object (false positive) $\mathbf{e}$ is registered into a database.
Let the set $\mathcal{E}=\{\mathbf{e}_1,\mathbf{e}_2,\cdots\}$ denote all registered error items after sufficient simulation time, and scenarios containing false detections are deemed important.
The cardinality of the set $\{\mathbf{e}\in\mathcal{E}:\|\mathbf{e}-\mathbf{x}\|\leq r\}$ represents the expected number of false detections that could be calibrated by the road sensor, which is to be maximized, where $\mathbf{x}$ and $r$ are the position and accurate-detection range of the road sensor.
The problem of optimizing $\mathbf{x}$ is a discrete optimization problem where a finite set of possible locations is available, as sensors can only be attached to utility poles.
In addition, $r$ is a monotonically increasing function of the infrastructure cost.
For example, in object detection task, $r$ increases with the number of laser channels, but this will also increase the LiDAR price.
The VRCSP method can be adopted to determine the positions of multiple sensor suites $(\mathbf{x}_1,\mathbf{x}_2,\cdots)$.
The associated result is shown in Fig.~4, where we have placed 2 road sensor suites at the cross-road (i.e., positions 1 and 4) and 1 sensor suite at the straight-road (i.e., position 8).
The perception range of the roadside infrastructure is significantly improved by adopting the multi-modal multi-view sensor fusion techniques.
The road sensors are connected via optical fiber so that the associated transmission delay can be ignored.

\begin{figure*}[!t]
\centering
\subfigure[]{\includegraphics[width=1\textwidth]{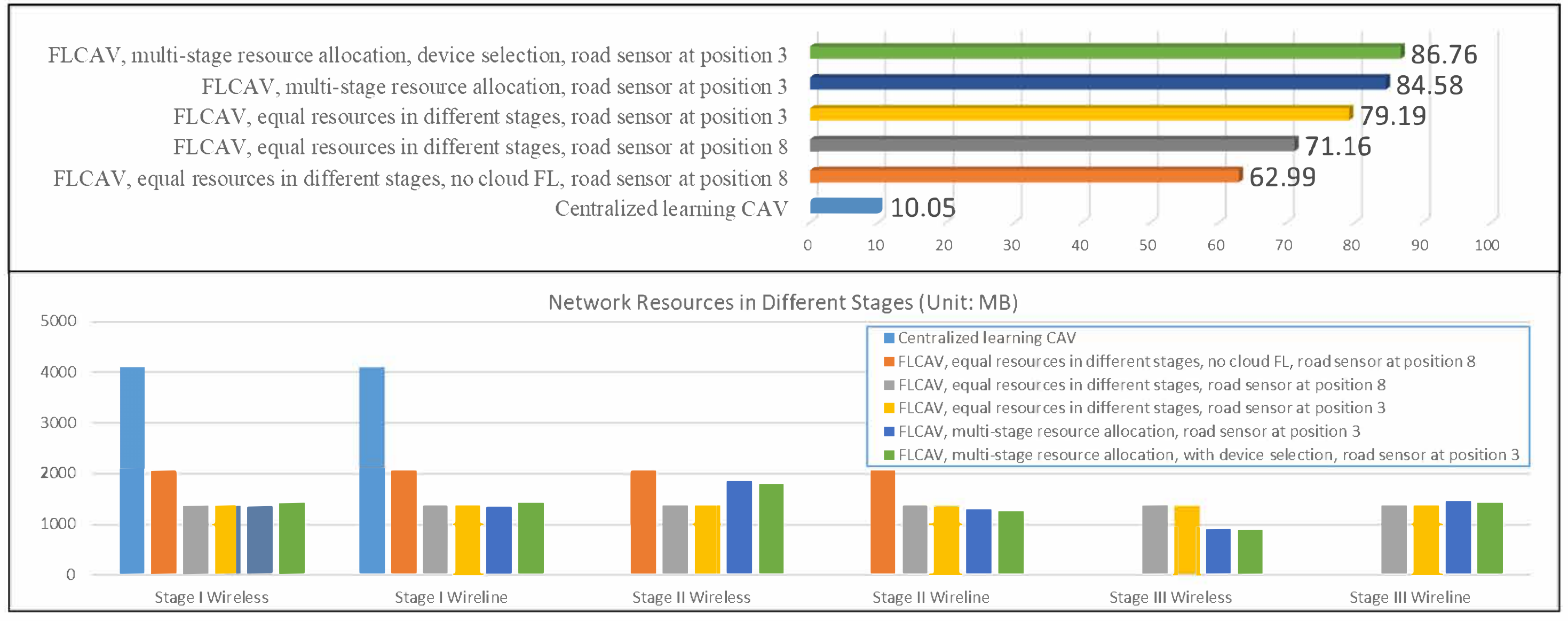}}
\subfigure[]{\includegraphics[width=1\textwidth]{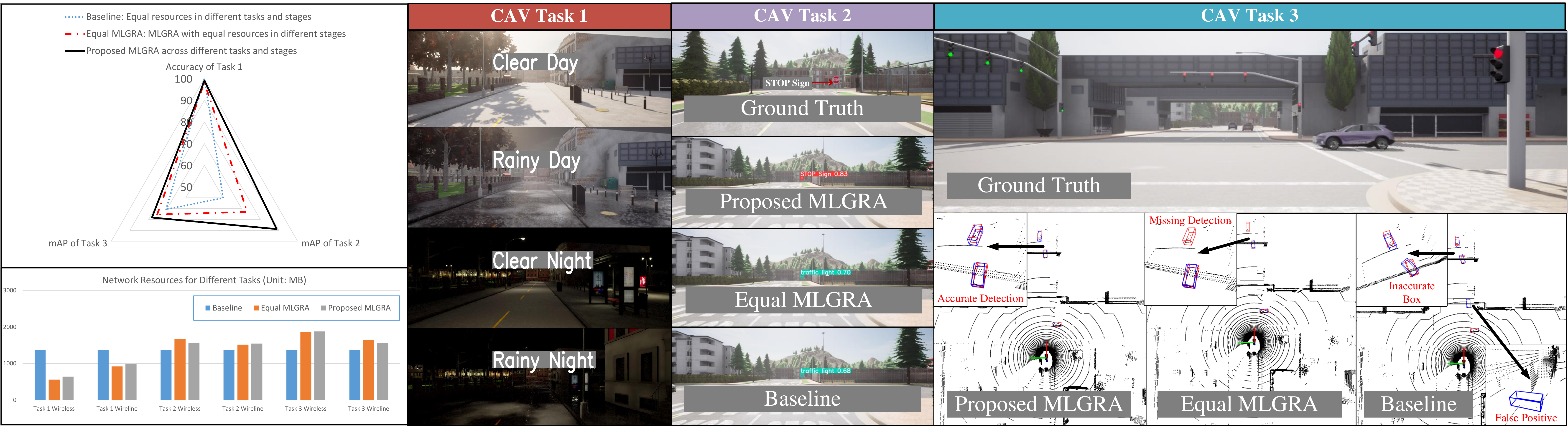}}
\caption{{\color{black}a) Comparison among different schemes for single-task perception.
Dataset: 5000 point-cloud samples in Town02 for pretraining; 4 CAVs (each with 500 samples) in Town05 for edge FL; 3 CAVs (each with 500 samples) in Town03 for cloud FL; 2432 samples in Town05 for testing.
b) Comparison among different schemes for multi-task perception.
Task-1 dataset: 4000 RGB images in Town02 for pretraining dataset; 4 CAVs (each with 100 images) in Town05 for edge FL; 3 CAVs (each with 100 images) in Town03 for cloud FL; 2000 images in Town02 for testing.
Task-2 dataset: 2500 RGB images in Town02 for pretraining; 4 CAVs (each with 1000 images) in Town05 for edge FL; 3 CAVs (each with 1000 images) in Town03 for cloud FL; 3279 images in Town05 for testing.
Task-3 dataset: same as Fig.~6a with road sensor placed at position 3.}
}
\end{figure*}

\section{Implementation and Experiments}

\subsection{Software Architecture}

CarlaFLCAV (shown in Fig.~5a) is an open-source FLCAV simulation platform that supports:
(1) multi-modal dataset generation, including point-cloud, image, radar data with associated calibration, synchronization, and annotation;
(2) training and inference examples for CAV perception, including weather classification, traffic sign detection, object detection, and trajectory prediction; (3) various FL frameworks, including FedAvg, device selection, noisy aggregation, parameter selection, distillation, and personalization; and
(4) optimization modules, including network resource and road sensor placement optimization.
The implementation of (2) is based on LeNet-5, Yolov5, OpenPCDet, LSTM, with the associated results shown in Fig.~1b. The implementation of (4) is illustrated in Fig.~3.
Below we focus on (1) and (3).

Specifically, raw sensory data is recorded using CARLA \cite{carla}, which adopts Unreal Engine 4 for state-of-the-art visual rendering and physics simulation.
Then, calibration represents sensed information in a common coordinate system via rotation and transition matrices.
CarlaFLCAV assumes perfect calibration, but practical systems may involve errors since sensors' intrinsic (e.g., shape of the camera lens) and extrinsic (i.e., pose) parameters need to be estimated.
Time stamping adopts LiDAR as a reference, i.e., each laser spin is a frame.
Synchronization among different sensors can be realized via hardware or software trigger, and the worst-case time difference is at the millisecond level. Finally, data annotation tracks the categories, poses, and occlusions of objects via CARLA APIs (\href{https://carla.readthedocs.io/en/latest/}{https://carla.readthedocs.io/en/latest/}).

For wireline FL, the DNN parameter exchange can be implemented based on the Robot Operating System (ROS) communication, which offers inter-process communication among distributed nodes by publishing/subscribing topics.
For wireless FL, due to limited capacity of wireless channels, only a subset of CAVs can be selected to convey their parameter updates at each FL round.
Thus, CarlaFLCAV implements importance-aware device selection, where vehicles with larger gradient norms are assigned a higher probability to be scheduled.
Besides, CarlaFLCAV includes noisy aggregation, which injects random noises into the DNN parameters as a mask to protect the data privacy against model inversion attacks.
DNN frozen is optional, which fixes partial layer parameters to reduce the communication cost.

\subsection{Experimental Validation}

First, to evaluate how close CarlaFLCAV is to real-life conditions, we train the SECOND (in Fig.~1b) with 3000 point cloud samples generated by CarlaFLCAV for 60 epochs and test the trained SECOND on a real-world dataset KITTI. The mean average precision (mAP) at IoU$=0.5$ is $58\%$ for object detection.
Qualitative results are shown in Fig.~5b, where the DNN trained with CarlaFLCAV detects all objects in real data.

Next, to verify the effectiveness of the multi-stage FLCAV, we consider the single-task case (i.e., task 3 in Fig.~1b) and compare the mAP of final DNNs for different schemes, under the same wireless and wireline uplink resource budgets (i.e., $4\,$GB). The settings and results are shown in Fig.~6a.
Our findings are summarized below.
\begin{itemize}
\item[(i)]
All FL schemes achieve higher mAPs than centralized learning.
This demonstrates the necessity of exploiting domain-information of Town05.

\item[(ii)]
With cloud FL, the mAP is improved, since the CAVs in Town03 provide new knowledge about corner-case and occluded objects.

\item[(iii)]
With equal resources in different stages, the VRCSP placing the road sensor at position 3 significantly improves the mAP compared with the conventional method placing the sensor at position 8.

\item[(iv)]
By jointly optimizing the network resources across three stages, the mAP is further increased to 84.58, which demonstrates the effectiveness of MLGRA.
Our result implies that more wireless (wireline) resources should be allocated to the edge (cloud) FL stage.

\item[(v)]
Under the same VRCSP and MLGRA methods,
the mAP score with device selection is slightly higher than direct FL, as the stragglers are removed from FL groups.
\end{itemize}

Finally, we simulate the multi-task multi-modal perception, including tasks 1--3 in Fig.~1b. The settings and results are shown in Fig.~6b.
Our findings are summarized below.
\begin{itemize}
\item[(i)] Since our goal is to maximize the perception accuracy of all tasks, a larger area indicates better performance. The baseline scheme (i.e., blue dotted-line) has the smallest area, which can be treated as a worst-case performance bound.

\item[(ii)] Leveraging the performance predictor in Fig.~3a, the equal MLGRA method (i.e., red dashed-line) achieves a larger triangle area than that of the baseline.
This is because it automatically allocates more resources to object detection and sign recognition as shown in Fig.~6b, which are harder tasks.

\item[(iii)]
With joint resource allocation across different stages and tasks, the proposed MLGRA method (i.e., black solid-line) achieves the largest triangle in Fig.~6b. The perception accuracies are $99.0, 88.8, 78.16$ for tasks $1, 2, 3$, respectively.

\item[(iv)]
For task 2, the proposed MLGRA successfully detects the STOP sign in cloudy days, while other methods misclassify the STOP sign as a traffic light.
For task 3, the proposed MLGRA successfully detects all the objects at the cross-road, while other methods involve inaccurate detections and false positives.
\end{itemize}

\section{Conclusion}

This article has reviewed the integration of FL and CAV for overcoming perception challenges in open driving scenarios.
The multi-tier networking, multi-stage training, and multi-task generation frameworks of FLCAV were presented.
The MLGRA and VRCSP methods were proposed to solve the network management and sensor placement problems, respectively.
The frameworks and methods were verified in a software platform CarlaFLCAV.
Future directions are listed below.

\textbf{Simulation-to-reality transfer}.
Real-world CAV datasets involve far more physical mechanisms (e.g., illuminations) and interactive behaviors (e.g., competitions) than simulation datasets.
Hence, there exists a non-negligible gap between the simulation and the reality, making transfer learning and real-world testing indispensable for FLCAV.

\textbf{Autonomous driving under perception uncertainties}.
Perception uncertainties will propagate to the subsequent planning system.
Hence, scientific approaches for computing the perception uncertainties and adjusting the safety distance in collision avoidance constraints are needed. End-to-end autonomous driving that directly maps sensor inputs into vehicle actions is also a promising solution to address the uncertainty propagation issue.

\section{Acknowledgement}

This work was supported by Science and Technology Development Fund of Macao S.A.R (FDCT) (No. 0015/2019/AKP), National Natural Science Foundation of China (No. 62001203), Shenzhen Science and Technology Program (No. RCB20200714114956153, JCYJ20200109141622964), Intel ICRI-IACV Research Fund (CG\#52514373), and Guangdong Basic and Applied Basic Research Project (No. 2021B1515120067). This work was also supported by Australian Research Council's Discovery Project (DP210102169).

\section*{Biography}

\textbf{Shuai Wang} [M'19] (s.wang@siat.ac.cn) received the Ph.D. degree from the University of Hong Kong in 2018. He is currently an Associate Professor at the Shenzhen Institute of Advanced Technology, Chinese Academy of Sciences. His research interests include autonomous systems and communications. He received the Best Paper Awards from IEEE ICC in 2020 and IEEE SPCC in 2021.

\vspace{0.15in}

\textbf{Chengyang Li} [S'22] (cli386@connect.hkust-gz.edu.cn) received a bachelor's degree in computer science from the Southern University of Science and Technology. He is currently a Master of Philosophy (M.Phil.) student at the Hong Kong University of Science and Technology (Guangzhou). His research interest includes robot localization, mapping, and motion planning.

\vspace{0.15in}

\textbf{Derrick~Wing~Kwan~Ng} [F'21] (w.k.ng@unsw.edu.au) received a bachelor's degree with first-class honors and a Master of Philosophy (M.Phil.) degree in electronic engineering from the Hong Kong University of Science and Technology (HKUST) in 2006 and 2008, respectively. He received his Ph.D. degree from the University of British Columbia (UBC) in Nov. 2012. He was a senior postdoctoral fellow at the Institute for Digital Communications, Friedrich-Alexander-University Erlangen-N\"urnberg (FAU), Germany. He is now working as a Scientia Associate Professor at the University of New South Wales, Sydney, Australia.  His research interests include global optimization, physical layer security, IRS-assisted communication, UAV-assisted communication, wireless information and power transfer, and green (energy-efficient) wireless communications.

\vspace{0.15in}

\textbf{Yonina C. Eldar} [F'12] (yonina.eldar@weizmann.ac.il) is a professor in the Department of Math and Computer Science, Weizmann Institute of Science, where she heads the Center for Biomedical Engineering and Signal Processing. She is a fellow of IEEE, a fellow of EURASIP, and a member of the Israel Academy of Sciences and Humanities.

\vspace{0.15in}

\textbf{H. Vincent Poor} [F'87] (poor@princeton.edu) is the Michael Henry Strater University Professor at Princeton University, where his interests include wireless networks, energy systems and related fields. Among his publications is the recent book \emph{Machine Learning and Wireless Communications} (Cambridge University Press, 2022). He received the IEEE Alexander Graham Bell Medal in 2017.

\vspace{0.15in}

\textbf{Qi Hao} [M'05] (hao.q@sustech.edu.cn) received the Ph.D. degree in electrical and computer engineering from Duke University in 2006. He is currently an Associate Professor with the Department of Computer Science and Engineering, Southern University of Science and Technology (SUSTech), Shenzhen 518055, China. He is also with Shenzhen Research Institute of Trustworthy Autonomous Systems.

\vspace{0.15in}

\textbf{Chengzhong Xu} [F'16] (czxu@um.edu.mo) is currently a Chair Professor of Computer Science in University of Macau. His research interests are in cloud and distributed computing, systems support for AI, smart city and autonomous driving.  He published two research monographs and more than 400 journal and conference papers. Dr. Xu was the Chair of IEEE Technical Committee on Distributed Processing from 2015 to 2020.  He was a recipient of the Faculty Research Award, Career Development Chair Award, and the President's Award for Excellence in Teaching of WSU. He was also a recipient of the ``Outstanding Oversea Scholar'' award of NSFC.


\begin{thebibliography}{15}

\bibitem{perception} D.~Feng et al., ``Deep multi-modal object detection and semantic segmentation for autonomous driving: Datasets, methods, and challenges,'' \emph{IEEE Trans. Intell. Transp. Syst.}, vol.~22, no.~3, pp.~1341--1360, Mar.~2021.

\bibitem{flcav_app1} S.~Savazzi, M.~Nicoli, M.~Bennis, S.~Kianoush, and L.~Barbieri, ``Opportunities of federated learning in connected, cooperative and automated industrial systems,'' \emph{IEEE Commun. Mag.}, vol.~59, no.~2, pp.~16--21, Feb.~2021.

\bibitem{flcav_app2} B.~Liu, L.~Wang, M. Liu, and C.~Xu, ``Federated imitation learning: A novel framework for cloud robotic systems with heterogeneous sensor data,'' \emph{IEEE Robot. Autom. Lett.}, vol.~5, no.~2, pp.~3509--3516, Apr.~2020.

\bibitem{flcav_netw1} J.~Posner, L.~Tseng, M.~Aloqaily, and Y.~Jararweh, ``Federated learning in vehicular networks: Opportunities and solutions,'' \emph{IEEE Netw.}, vol.~35, no.~2, pp.~152--159, Mar./Apr. 2021.

\bibitem{flcav_netw2} J.~Zhang and K.~B.~Letaief, ``Mobile edge intelligence and computing for the Internet of vehicles,'' \emph{Proc. IEEE}, vol.~108, no.~2, pp.~246--261, Feb.~2020.

\bibitem{flcav_road1} E.~Arnold, M.~Dianati, R.~de Temple, and S.~Fallah, ``Cooperative perception for 3D object detection in driving scenarios using infrastructure sensors,'' \emph{IEEE Trans. Intell. Transp. Syst.}, vol.~23, no.~3, pp.~1852--1864, Mar.~2022.

\bibitem{flcav_road2} Y.~Li, S.~Ren, P.~Wu, S.~Chen, C.~Feng, and W.~Zhang, ``Learning distilled collaboration graph for multi-agent perception,'' \emph{Adv. Neural Inf. Process. Syst.}, virtual, Dec.~2021, pp.~1--12.

\bibitem{cav_netw} I.~Yaqoob, L.~U.~Khan, S.~M.~A.~Kazmi, M.~Imran, N.~Guizani, and C.~S.~Hong, ``Autonomous driving cars in smart cities: Recent advances, requirements, and challenges,'' \emph{IEEE Netw.}, vol.~34, no.~1, pp.~174--181, Jan./Feb. 2020.

\bibitem{aads} W. Li et al., ``AADS: Augmented autonomous driving simulation using data-driven algorithms,'' \emph{Sci. Robot.}, vol.~14, no.~28, pp.~1--12, Mar. 2019.

\bibitem{carla} A. Dosovitskiy, G. Ros, F. Codevilla, A. Lopez, and V. Koltun, ``CARLA: An open urban driving simulator,'' in \emph{CoRL}, 2017, pp. 1--16.

\bibitem{flres1} M.~Chen, D.~G\"{u}nd\"{u}z, K.~Huang, W.~Saad, M.~Bennis, A.~V.~Feljan, and H.~V.~Poor, ``Distributed learning in wireless networks: Recent progress and future challenges,'' \emph{IEEE J. Sel. Areas Commun.}, vol.~39, no.~12, pp.~3579--3605, Dec. 2021.

\bibitem{personalization} Q.~Wu, X.~Chen, Z.~Zhou, and J.~Zhang, ``FedHome: Cloud-edge based personalized federated learning for in-home health monitoring,'' \emph{IEEE Trans. Mob. Comput.}, vol.~21, no.~8, pp.~2818--2832, Aug.~2022.

\bibitem{cav} A.~Eskandarian, C.~Wu, and C.~Sun, ``Research advances and challenges of autonomous and connected ground vehicles,'' \emph{IEEE Trans. Intell. Transp. Syst.}, vol.~22, no.~2, pp.~683--711, Feb.~2021.

\bibitem{task} Z.~Zheng, Q.~Chen, C.~Hu, D.~Wang, and F.~ Liu, ``On-edge multi-task transfer learning: Model and practice with data-driven task allocation,'' \emph{IEEE Tran. Parall. Distr.}, vol.~31, no.~6, pp.~1357--1371, Jun.~2020.

\bibitem{scenarios} X.~Zhang et al., ``Finding critical scenarios for automated driving systems: A systematic mapping study,'' \emph{IEEE Trans. Softw. Eng.}, early access, 2022. DOI: 10.1109/TSE.2022.3170122.


\end{thebibliography}
\end{document}